\documentclass[sigconf]{acmart}
\usepackage{MnSymbol}
\usepackage{enumitem}

\AtBeginDocument{%
  \providecommand\BibTeX{{%
    \normalfont B\kern-0.5em{\scshape i\kern-0.25em b}\kern-0.8em\TeX}}}

\copyrightyear{2025}
\acmYear{2025}
\setcopyright{rightsretained}
\acmConference[WSDM '25]{Proceedings of the Eighteenth ACM International Conference on Web Search and Data Mining}{March 10--14, 2025}{Hannover, Germany}
\acmBooktitle{Proceedings of the Eighteenth ACM International Conference on Web Search and Data Mining (WSDM '25), March 10--14, 2025, Hannover, Germany}\acmDOI{10.1145/3701551.3703525}
\acmISBN{979-8-4007-1329-3/25/03}

\begin{document}

\title{Graph Disentangle Causal Model: Enhancing Causal Inference in Networked Observational Data}

\author{Binbin Hu}
\authornote{Both authors contributed equally to this research.}
\email{bin.hbb@antfin.com}
\affiliation{%
  \institution{Ant Group}
  \country{China}
  \city{Hangzhou}
}

\author{Zhicheng An}
\authornotemark[1]
\email{1272002241@qq.com}
\affiliation{%
  \institution{Ant Group}
  \country{China}
  \city{Hangzhou}
}

\author{Zhengwei Wu}
\email{zhengwei_wu@yeah.net}
\affiliation{%
  \institution{Ant Group}
  \country{China}
  \city{Hangzhou}
}

\author{Ke Tu}
\email{tuke1993@gmail.com}
\affiliation{%
  \institution{Ant Group}
  \country{China}
  \city{Hangzhou}
}

\author{Ziqi Liu}
\email{ziqiliu@antgroup.com	}
\affiliation{%
  \institution{Ant Group}
  \country{China}
  \city{Hangzhou}
}

\author{Zhiqiang Zhang}
\email{ziqiliu@antgroup.com	}
\affiliation{%
  \institution{Ant Group}
  \country{China}
  \city{Hangzhou}
}

\author{Jun Zhou}
\email{ziqiliu@antgroup.com	}
\affiliation{%
  \institution{Ant Group}
  \country{China}
  \city{Hangzhou}
}

\author{Yufei Feng}
\email{fyf649435349@gmail.com}
\affiliation{%
  \institution{Alibaba Group}
  \country{China}
  \city{Hangzhou}
}

\author{Jiawei Chen}
\authornote{Corresponding author.}
\email{sleepyhunt@zju.edu.cn}
\affiliation{%
  \institution{Zhejiang University}
  \country{China}
  \city{Hangzhou}
}

\renewcommand{\shortauthors}{Binbin Hu et al.}



\begin{abstract}
Estimating individual treatment effects (ITE) from observational data is a critical task across various domains.
However, many existing works on ITE estimation overlook the influence of hidden confounders, which remain unobserved at the individual unit level.
To address this limitation, researchers have utilized graph neural networks to aggregate neighbors' features to capture the hidden confounders and mitigate confounding bias by minimizing the discrepancy of confounder representations between the treated and control groups.
Despite the success of these approaches, practical scenarios often treat all features as confounders and involve substantial differences in feature distributions between the treated and control groups.
Confusing the adjustment and confounder and enforcing strict balance on the confounder representations could potentially undermine the effectiveness of outcome prediction. 
To mitigate this issue, we propose a novel framework called the \textit{Graph Disentangle Causal model} (GDC) to conduct ITE estimation in the network setting.
GDC utilizes a causal disentangle module to separate unit features into adjustment and confounder representations. 
Then we design a graph aggregation module consisting of three distinct graph aggregators to obtain adjustment, confounder, and counterfactual confounder representations. 
Finally, a causal constraint module is employed to enforce the disentangled representations as true causal factors.
The effectiveness of our proposed method is demonstrated by conducting comprehensive experiments on two networked datasets.
\end{abstract}





\maketitle

\section{Introduction}


\begin{figure}
    \centering
    \includegraphics[width=0.5\textwidth]{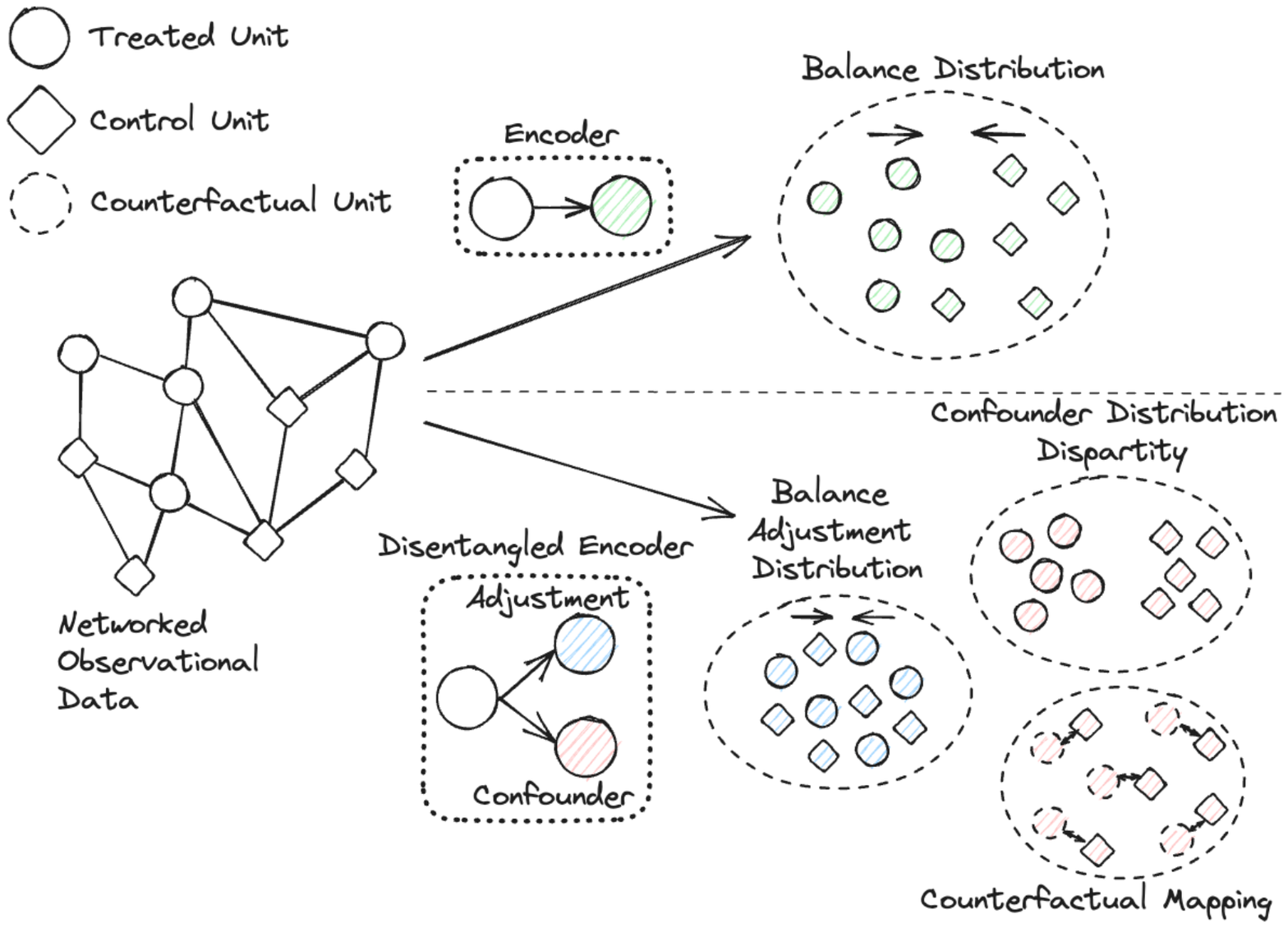}
    \caption{An illustration comparing conventional ITE estimation methods on networked data with our proposed approach. 
    Upper right: Conventional methods treat all features as confounder without identifying causal factors. Strict distribution balance may undermine outcome prediction efficacy.
    Lower right: Upon disentangling causal factors, we find that adjustment distributions between two groups can be effectively balanced, while confounder distributions cannot. Instead of enforcing distribution balance, we introduce a counterfactual mapping process to enhance ITE estimation.
    }
    \label{fig:intro_figure}
\end{figure}

Estimating individual treatment effects (ITE) has garnered significant attention and finds applications across diverse domains. In the healthcare sector, medical institutions strive to personalize treatments based on individual patient characteristics, aiming to optimize patient outcomes~\cite{glass2013causal}. In the realm of E-commerce, companies seek to predict the impact of item exposure on user engagement and satisfaction, enhancing their ability to deliver tailored recommendations~\cite{zhang2021causally}. 

However, many existing works on ITE estimation~\cite{hill2011bayesian,DBLP:conf/icml/ShalitJS17,wager2018estimation} overlook the influence of hidden confounders, which are the unobserved variables that causally affect both the treatment and the outcome. In essence, these methods heavily rely on the unconfoundedness assumption~\cite{morgan2015counterfactuals}, assuming that all confounders can be measured from the observed features and are sufficient to identify the treatment assignment mechanism. Nevertheless, without appropriately incorporating the influence of hidden confounders, these methods may yield a biased estimation.
To capture the presence of hidden confounders, recent studies have explored the use of network information, such as social network or patient similarity, in addition to traditional i.i.d observational data, in order to achieve more accurate ITE estimation~\cite{DBLP:conf/wsdm/GuoLL20,DBLP:conf/kdd/ChuR021,DBLP:conf/cikm/0002S22}. These approaches primarily utilize graph neural networks~\cite{DBLP:journals/tai/00010YAWP021} to encode the representations of target units by aggregating information from their neighboring units.

Despite their success, these methods treat all features as confounders without precisely identifing the latent factors present in networked data. 
To illustrate, let's consider the context of medical treatment. The patient's economic status serves as a confounding variable as it affects both access to expensive medicines and the patient’s rate of recovery. Conversely, variables like pre-existing health conditions and social support operate as adjustment variables, singularly affecting the patient's recovery rate. Notably, factors like economic status and social support attain comprehensive observability within the network context, such as a social network.
Moreover, these methods attempt to mitigate confounding bias by minimizing the discrepancy of confounder representations between the treated and control groups. However, in practical scenarios, there often exists substantial differences in the feature distributions between these groups. Returning to our medical example, 
enforcing strict balance on these representations of patients using expensive medicines and those without access could potentially undermine the effectiveness of outcome prediction.

To tackle the aforementioned challenges, we propose a novel ITE estimation framework called Graph Disentangle Causal Model (GDC), as described in Figure~\ref{fig:intro_figure}. Inspired by the great success of disentangled modeling in various applications~\cite{an2024ddcdr,gan2023matters}, our model aims to disentangle causal factors from networked observational data and utilize these disentangled factors to improve outcome prediction. Our proposed framework comprises three key components. 
Firstly, we introduce a causal disentangle module equipped with feature-wise mask to separate each unit features into adjustment and confounder factors. 
Secondly, we develop a graph aggregation module that incorporates of three distinct graph aggregators to generate embeddings for adjustment factors, confounding factors, and counterfactual confounding factors. 
The calculation of aggregation attention weights is based on the adjustment factors, as they exhibit unbiasedness with respect to treatment assignment and possess the ability to accurately measure neighbor similarity without introducing any bias.
Inspired from classic matching methods and leveraging the homophily characteristic of networked data, we generate the counterfactual confounding factors for each node by aggregating the confounder of neighbors in the opposite treatment group. This approach enables us to capture the influence of the confounders on the counterfactual outcomes. 
Finally, we employ a causal constraint module to comprise multiple loss functions, including adjustment distribution balance loss, treatment prediction loss, confounder mapping loss, and outcome predictions loss. These losses are jointly optimized under a multi-task training strategy to ensure the disentangled representations align with the true causal factors. Our main contribution can be summarized as,
\begin{itemize}[leftmargin=*]
    \item[$\blacktriangleright$] We investigate the problem of learning causal effects from networked observational data and emphasize the importance of disentangling node features to fully delineate the causal relationship between components on the graph.
    \item[$\blacktriangleright$] We propose a novel framework, Graph Disentangle Causal Model (GDC), which disentangles the origin features of each unit into two independent factors and designs a more targeted aggregation of different factors to achieve a full exploitation of causal relationships and control the confounding bias without damaging the predictive power of outcomes.
    \item[$\blacktriangleright$] We conduct comprehensive experiments on two semi-synthetic datasets compared with state-of-the-art methods and the results show our method achieves a significant improvement in terms of the causal metrics.
\end{itemize}



\section{Preliminary}
This section outlines technical preliminaries to ensure a common understanding of the concepts and terminology used in the subsequent discussion. In this work, scalars, vectors, and matrices are denoted by lowercase letters, boldface lowercase letters, and boldface uppercase letters, respectively. By default, $\mathbf{X}_{a,i}$ represents the $i$-th row of matrix $\mathbf{X}_{a}$, and $\mathbf{X}_{ij}$ denotes the $i$-th row and $j$-th column of the matrix $\mathbf{X}$.
Next, we formally present the problem statement of learning causal effects from networked observational data.

\textbf{Networked observational data.} The networked observational data is represented by $\mathcal{G} = <\mathbf{A}, \mathbf{X}, \mathbf{T}, \mathbf{Y}>$. $\mathbf{A} \in \mathbb{R}^{N \times N}$ represents the adjacency matrix with $N$ units. If there is an adjacency relationship between two units $v_{i}$ and $v_{j}$, we set $A_{ij} = 1$, and $A_{ij} = 0$ otherwise. $\mathbf{X} \in \mathbb{R}^{N \times K}$ denotes the feature matrix of units, where $K$ is the feature dimension. $\mathbf{T} = [t_{1}, \cdots, t_{N}]$ denotes the treatments where $t_{i} \in \{0,1\}$ represents the treatment received by unit $i$. In this paper, we focus on the binary treatment (e.g., $t_{i}=1$ indicates a treated unit while $t_{i}=0$ represents a control unit). $\mathbf{Y} = [y_{1}, \cdots, y_{N}]$ is the outcomes where $y_{i} \in \mathbb{R}$ is the continuous scalar outcome of unit $i$ when receives treatment $t_{i}$.

\textbf{Learning Causal Effects.} We adopt the potential outcome framework to estimate treatment effects. The potential outcome $y_{i}^{t}$ is defined as the value of outcome would have taken if the treatment of unit $i$ had been set to $t$. Then the individual treatment effects (ITE) can be formally defined as the difference between the expected potential treated and control outcomes,
\begin{equation}
    \tau_{i} = \tau(\mathbf{X}_{i},\mathbf{A}) = \mathbb{E}[y_{i}^{1}|\mathbf{X}_{i},\mathbf{A}]-\mathbb{E}[y_{i}^{0}|\mathbf{X}_{i},\mathbf{A}].
\end{equation}
ITE represents the causal effects on the improvement of the outcome resulting from the treatment. We further define the average treatment effects (ATE) as the average of ITE over all units as $ATE = \frac{1}{N} \sum_{i=1}^{N}\tau_{i}$. The objective of this paper is to learn a function $\mathcal{F}: \mathcal{F}(\mathbf{X}, \mathbf{A}, \mathbf{T}) \rightarrow \mathbf{Y}$ that utilizes the features, network structure and treatments to predict the potential outcomes and estimate the ITE and ATE in the networked observational data. Since we can only observe at most one potential outcome from the observational data, known as the factual outcome, the main challenge lies in inferring the counterfactual outcome $y_{i}^{CF} = y_{i}^{1-t_{i}}$.

We define the unconfoundedness assumption and confounding bias as follows:

\textbf{Unconfoundedness.} Conditional to the features $\mathbf{x}$ and the adjacency matrix $A$, the potential outcomes are independent of treatment assignment $t$, i.e., $(y^{1}, y^{0}) \upmodels t | \mathbf{x}, A$. Traditional causal inference tasks only consider the effect of features while neglecting the network effect, i.e., $(y^{1}, y^{0}) \upmodels t | \mathbf{x}$.

\textbf{Confounding bias.} For any $t \in \{0, 1\}$, the distribution of confounder representation $p(\mathbf{E}_{c}|T=t)$ on treatment $T=t$ differs from the 
 counterfactual distribution of confounder representation $p(\mathbf{E}_{cf}|T\neq t)$ on treatment $T\neq t$. In such a situation, the counterfactual prediction is biased.






\begin{figure}
    \centering
    \includegraphics[width=0.5\textwidth]{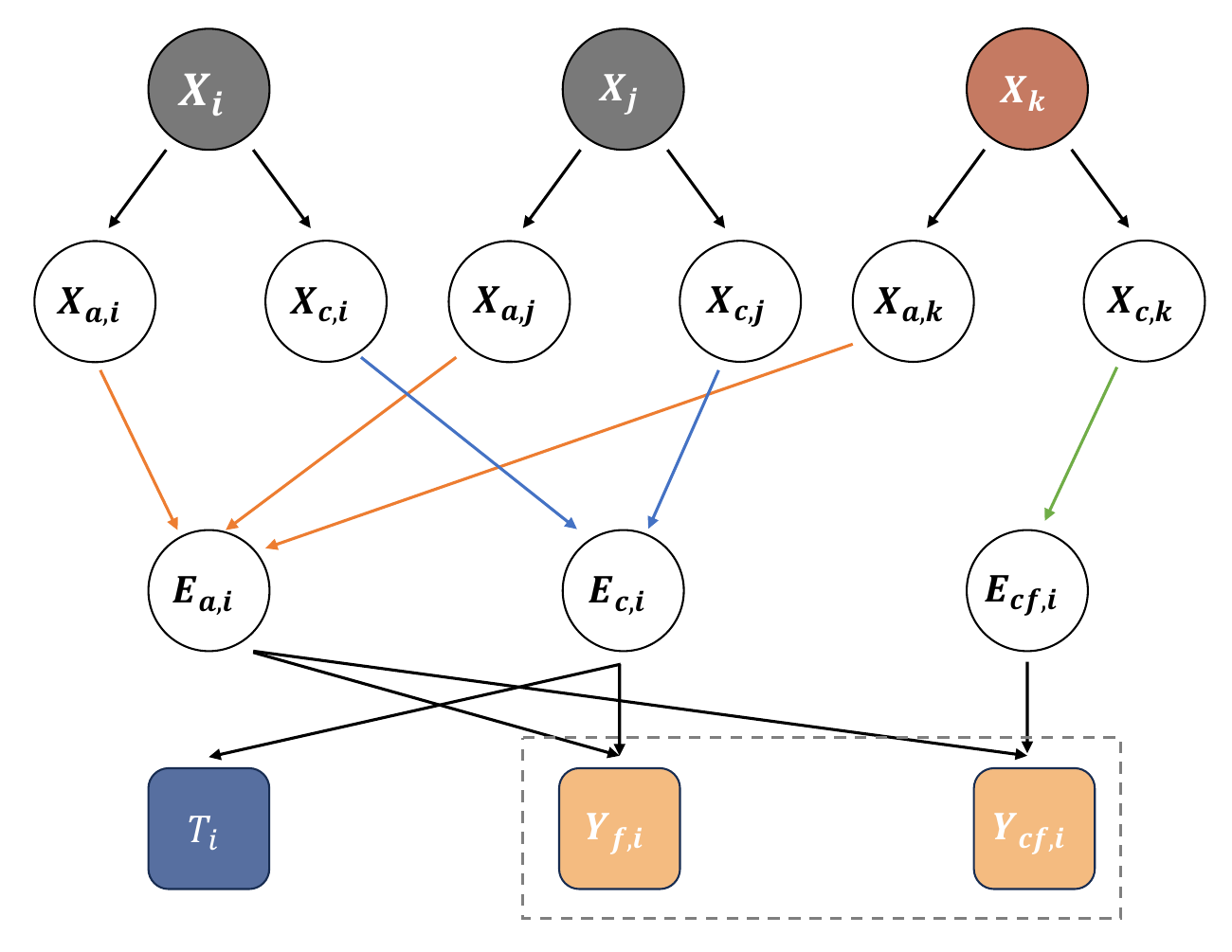}
    \caption{Causal graph for networked observational data. $\mathbf{X}_i, \mathbf{X}_j, \mathbf{X}_k$ represent the features of unit $i$, neighboring unit $j$ with the same treatment as unit $i$, and neighboring unit $k$ with a different treatment, respectively. $\mathbf{X}_{a, \cdot}$, $\mathbf{X}_{c, \cdot}$ are their corresponding disentangled adjustment and confounder from their own features. And $\mathbf{E}_{a, i}, \mathbf{E}_{c, i}, \mathbf{E}_{cf, i}$ are the adjustment, confounder, and counterfactual confounder by incorporating network information. $\mathbf{Y}_{f,i}, \mathbf{Y}_{cf,i}$ are factual outcome and counterfactual outcome.}
    \label{fig: causal_graph}
\end{figure}

\section{Methodology}
\begin{figure*}
    \centering
    \includegraphics[width=1.0\textwidth]{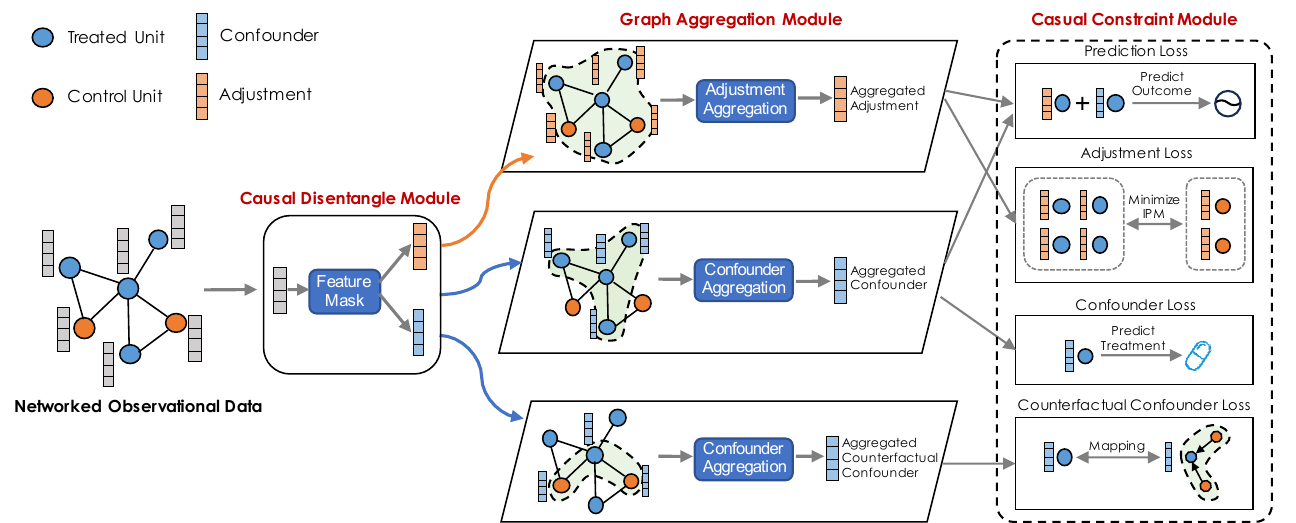}
    \caption{Overall architecture of our proposed Graph Disentangle Causal Model.}
    \label{fig: main_structure}
\end{figure*}

In this section, we introduce a novel framework, the Graph Disentangle Causal Model, to conduct ITE estimation.
We depict the causal graph of causal inference on networked observational data in Figure \ref{fig: causal_graph}.
In the causal graph, each node is first disentangled into adjustment and confounder parts based on their own features.
However, due to the network effect, the adjustment and confounder representations based on their own features are incomplete.
In order to consider the influence of their neighbors, more accurate adjustment and confounder representations need to be obtained. 
Additionally, the adjustment and confounder representations are influenced differently by neighbors, and those with opposite treatment provide good approximate counterfactuals.
By incorporating network-enriched adjustment, confounder and counterfactual confounder representations, we can predict the true label and counterfactual results.
Based on this causal graph, we design our model, as shown in Figure \ref{fig: main_structure}.
In the first {\em Graph Disentangle Module}, we separate the features of each node into two distinct representations.
We expect these two representations to be adjustment and confounder representations, and we add constraints in the {\em Causal Constraint Module} to ensure that this goal is achieved.
To use neighbors with both the same treatment and opposite treatment, we use different neighbors for different factor aggregators.
Specifically, we aggregate neighbors' adjustment and confounder representations by using three distinct graph aggregators to generate embeddings for adjustment factors, confounding factors, and counterfactual confounding factors in the {\em Graph Aggregation Module}. 
In the last {\em Causal Constraint Module}, we add multiple constraints to ensure the representation learning follows the causal graph described above.

\subsection{Causal Disentangle Module}
\label{sec: causal_disentangle}

Traditional causal inference methods, including propensity score-based methods and representation balancing methods~\cite{rosenbaum1983central,austin2011introduction,hirano2003efficient,funk2011doubly,hainmueller2012entropy,imai2014covariate}, commonly treat all features as confounders and aim to mitigate confounding bias by minimizing the discrepancy of feature distribution between the treated and control groups. However, recent studies have shed light on the limitations of treating all features as a whole and have made notable advancements in disentangled representation learning methods for estimating treatment effects~\cite{DBLP:conf/aaai/KuangCLJY017,DBLP:conf/iclr/HassanpourG20,WuLearningEstimation}. 
By identifying disentangled factors of features and treating them differently, we can improve the accuracy of potential outcome prediction and reduce the negative influence of confounders on treatment effect estimation.
Following the previous work, we assume that the feature variable $\mathbf{X}$ can be decomposed into two kinds of latent variables: adjustment variables that only determine the outcome, and confounder variables that influence both the treatment and outcome.

To disentangle the features, we utilize a feature-wise mask \cite{DBLP:journals/corr/abs-2102-07619} by applying an instance-guided embedding feature mask which performs element-wise multiplication on feature embeddings. 
To generate the feature embedding, we employ two fully connected layers (FC) with non-linear activation functions. These layers are responsible for constructing the instance-guided mask, enabling the incorporation of global contextual information from input instances to dynamically emphasize informative elements in the feature embeddings. The output dimension of the second FC layer is set equal to the feature dimension. Notably, the sigmoid function is applied to the positive and negative outputs of the FC layers, generating two distinct masks for adjustment and confounder representations. The mask learning process can be denoted as

\begin{equation}
    \mathbf{Z} = (\operatorname{ReLu}(\mathbf{X}\mathbf{W}_{1} + \boldsymbol{\Theta}_{1}))\mathbf{W}_{2}+\boldsymbol{\Theta}_{2},
\end{equation}
\begin{equation}
    \mathbf{V}_{mask} = \sigma(\mathbf{Z}),
\end{equation}
\begin{equation}
    \mathbf{V}_{mask}^{\prime} = \sigma(\mathbf{-Z}),
\end{equation}
where $\mathbf{W}_{1} \in \mathbb{R}^{K \times D}$, $\mathbf{W}_{2} \in \mathbb{R}^{D \times K}$ are weighted parameters, $\boldsymbol{\Theta}_{1} \in \mathbb{R}^{D \times 1}$, $\boldsymbol{\Theta}_{2} \in \mathbb{R}^{K \times 1}$ are learned bias of the two FC layers, and $\sigma(\cdot)$ is the sigmoid function.
$K$ and $D$ are dimensions of feature embedding and hidden layer respectively. Then an element-wise multiplication is performed to incorporate the global information by instance-guided mask and generate the two masked features which denote confounder and adjustment:
\begin{equation}
    \mathbf{X}_{c} = \mathbf{V}_{mask} \odot \mathbf{X},
\end{equation}
\begin{equation}
    \mathbf{X}_{a} = \mathbf{V}_{mask}^{\prime} \odot \mathbf{X},
\end{equation}
where $\mathbf{X}_{c}$ and $\mathbf{X}_{a}$ denote confounder and adjustment representations. 
Then $\sigma(\mathbf{-Z_{i}}) = 1-\sigma(\mathbf{Z}_{i})$ is complementary for each unit $i$ and it leads to $\mathbf{X}_c+\mathbf{X}_a=\mathbf{X}$.
The instance-guided feature mask can be viewed as an attention mechanism that enables each disentangled component to focus on its most relevant features.
Nevertheless, the feature disentangle module cannot guarantee that the disentangled representations are confounder and adjustment representations.
To overcome this limitation, we must add constraints to achieve this goal.
We discuss these constraints in the Causal Constraint Module.

\subsection{Graph Aggregation Module}
It is not entirely accurate to estimate adjustment and confounder based solely on users' individual characteristics, as users' behaviors are often influenced by their friends, i.e., the network effect.
Consequently, our objective in this section is to capture the causal network effect between target units and their neighboring units.
To achieve this, we introduce a graph aggregation module. This module integrates the adjustment and confounder representations of the neighboring units, allowing us to generate the hidden unobserved causal factors of the target unit.
By incorporating the information from neighboring units, we are able to effectively capture the influence of these units on the outcome of the target units.
The influence of adjustment and confounder representations by their neighbors' adjustment and confounder representations conforms to different mechanisms owing to the characteristics of adjustment and confounder.
Therefore, we have designed different aggregators for each of them, as follows subsections.

\subsubsection{Adjustment Aggregation}
Since the adjustment variables remain unaffected by the treatment assignment and exhibit a balanced distribution between treated and control groups, we regard them as inherent features that reflect the fundamental characteristic of individual units.
In light of this, we aggregate the adjustments of neighboring units to obtain hidden adjustment representations of the target unit. Furthermore, we utilize the similarity of adjustments among neighbors as a stable measure of attention, enabling us to assess the influence of neighboring units on the target unit.
By aggregating the adjustment representations of neighbors through this stable attention mechanism, we obtain a more precise aggregated adjustment representation, as follows:
\begin{equation}
    \label{gat_equ}
    \mathbf{E}_{a,i} = \sigma\left(\sum_{j \in \mathcal{N}(i)} \alpha_{i j} \mathbf{W}_a\mathbf{X}_{a,j}\right),
\end{equation}
where $\mathcal{N}(i)$ is the set of neighbors of unit $i$, $\mathbf{W}_a$ is a shared linear transformation's weight matrix, and $\mathbf{E}_{a,j}$ is the adjustment representation of unit $i$.
$\alpha_{i j}$ denotes the attention coefficients computed by the self-attention mechanism among the adjustment representations of unit $i$ and unit $j$ which can be expressed as
\begin{equation}
   \label{aggeration_equ}
   \alpha_{i j}=\frac{\exp \left(\operatorname{LeakyReLU}\left(\mathbf{W}_{att}[\mathbf{E}_{a,i}||\mathbf{E}_{a,j}]\right)\right)}{\sum_{k \in \mathcal{N}(i)} \exp \left(\operatorname{LeakyReLU}\left(\mathbf{W}_{att}[\mathbf{E}_{a,i}||\mathbf{E}_{a,k}]\right)\right)},
\end{equation}
where $\mathbf{W}_{att}$ denotes the trainable parameters, $||$ represents the concatenate function and $\operatorname{LeakyReLU}$ is a non-linear activation.

\subsubsection{Confounder and Counterfactual Confounder Aggregation}
As the confounder affects treatment assignment, there are two types of neighbors for confounder aggregation: the neighbors of the same treatment with the target unit for factual confounder aggregation and the neighbors of different treatments with the target unit for counterfactual confounder aggregation.
Based on the treatment group, we split the graph into two subgraphs and aggregate over them respectively.
However, since the confounder is related to the treatment, it is not suitable to measure the similarity of units based on the distance of confounder representations.
Doing so would result in a significant difference between the similarity of unit pairs with the same treatment and those with different treatments.
Instead, we use the normalized attention calculated by the adjustment $\frac{\operatorname{exp}(\alpha_{i j})}{\sum_j\operatorname{exp}(\alpha_{i j})}$ in the previous subsection as the importance to aggregate neighbors, as the distance between adjustments is unrelated to the treatment but can measure stable relationships between units.
Using this approach, we obtain the confounder representations $\mathbf{E}_{c,i}$ and counterfactual confounder representations $\mathbf{E}_{cf,i}$ by two different aggregators as follows:
\begin{gather}
    \mathbf{E}_{c,i} = \sigma\left(\sum_{j \in \mathcal{N}(i),t_i=t_j} \frac{\operatorname{exp}(\alpha_{i j})}{\sum_{j\in \mathcal{N}(i),t_i=t_j}\operatorname{exp}(\alpha_{i j})} \mathbf{W}_c \mathbf{X}_{c,j}\right), \\
    \mathbf{E}_{cf,i} = \sigma\left(\sum_{j \in \mathcal{N}(i),t_i\neq t_j} \frac{\operatorname{exp}(\alpha_{i j})}{\sum_{j\in \mathcal{N}(i),t_i\neq t_j}\operatorname{exp}(\alpha_{i j})} \mathbf{W}_{cf} \mathbf{X}_{c,j}\right),
\end{gather}
where $\mathbf{W}_c$ and $\mathbf{W}_{cf}$ are the weight matrix of confounder and counterfactual confounder aggregation.

\subsection{Causal Constraint Module}
After the graph aggregation process, units have three generated representations: the aggregated adjustment embedding $\mathbf{E}_{a}$, the aggregated factual confounder embedding $\mathbf{E}_{c}$, and the aggregated counterfactual confounder embedding $\mathbf{E}_{cf}$. 
To ensure the disentangled representations as true causal factors and the algorithm process follows the causal graph in Figure ~\ref{fig: causal_graph}, we need to add causal constraints based on their individual characteristics.

Firstly, it is essential to impose appropriate constraints on the learned aggregated representation to ensure that they accurately correspond to the true adjustment and confounder.
For the adjustment variable, since it is independent of the treatment assignment, we minimize the discrepancy of adjustment between treated and control groups to achieve precise disentanglement of the adjustment factor as follows:
\begin{equation}
    \operatorname{disc}\left(\{\mathbf{E}_{a}\}_{i:t_i=0},\{\mathbf{E}_{a}\}_{i:t_i=1}\right).
\end{equation}
Many integral probability metrics (IPMs), such as Maximum Mean Discrepancy (MMD) \cite{DBLP:journals/jmlr/GrettonBRSS12} and Wasserstein distance \cite{DBLP:journals/corr/ArjovskyCB17}, can be used to measure the discrepancy of distributions. Without loss of generality, we adopt an efficient approximation version of Wasserstein-1 distance \cite{cuturi2014fast} as $\operatorname{disc}$ function in the previous equation. The adjustment loss of the task is defined as
\begin{equation}
    \mathcal{L}_{adjustment} = \operatorname{Wass}\left(\{\mathbf{E}_{a}\}_{i:t_i=0},\{\mathbf{E}_{a}\}_{j:t_j=1}\right),
\end{equation}
where $\operatorname{Wass}$ represents the Wasserstein-1 distance metric.

In contrast to the adjustment variables, the confounder variables can determine the assigned treatment.
Therefore, we introduce a task to maximize the predictive capability of the confounder for treatment assignment, which guides the disentanglement process of the confounder. This task resembles the approach employed in propensity score-based methods, but in our approach, we solely utilize the decomposed variable instead of the entire feature set. The loss for the confounder disentangle task can be expressed as:
\begin{equation}
    \mathcal{L}_{confouder} = \mathcal{L}(t,Pr(T=1|\mathbf{E}_{c})),
\end{equation}
where $\mathcal{L}_{confouder}$ denotes the binary cross-entropy loss metric and $Pr(\cdot)$ represents a classifier based on deep neural networks, which outputs the probability of receiving treatment.

By the two mentioned losses, the aggregated adjustment $\mathbf{E}_{a}$ and confounder representations $\mathbf{E}_{c}$ serve as approximations of the adjustment and confounder, respectively.
It is important to note that the adjustment representation is independent of the confounder representation.
Considering that the aggregated adjustment representations are a function of both the units' own adjustment representations $\mathbf{X}_{a, i}$ and their neighbors' adjustment representations $\{\mathbf{X}_{a, j} | j \in \mathcal{N}(i)\}$, it follows that the own adjustment representations $\mathbf{X}_{a, i}$ are also unrelated to the aggregated confounder representations.
Consequently, this implies that the own adjustment representations, $\mathbf{X}{a, i}$, should reflect the adjustment based on the units' own features, and the same applies to the own confounder representations $\mathbf{X}_{c, i}$.

For factual outcome prediction, we employ the combination of aggregated adjustment and confounder, while for counterfactual outcome prediction, we use the combination of aggregated adjustment and counterfactual confounder.
However, the homophily characteristic causes many units to have few or no neighbors under different treatments, which results in a lack of information for the aggregated counterfactual confounder. To address this issue, we learn a representation mapping function $g: g(\mathbf{X}_{c,i}, \mathbf{E}_{a,i}) \rightarrow \mathbf{E}_{cf,i}$ from self-confounder and aggregated adjustment to aggregated counterfactual confounder. 
This function establishes the connection between the self-confounder and the adjustment to its potential counterfactual confounder.
Additionally, we adopt the mean squared error as the loss metric for the confounder mapping function $g$ if there are neighbors with opposite treatment: 
\begin{equation}
    \mathcal{L}_{cf\text{-}confounder} = \frac{1}{N} \sum_{i=1}^N\left(g(\mathbf{X}_{c,i}, \mathbf{E}_{a,i}) - \mathbf{E}_{cf,i} \right)^2.
\end{equation}
In such a way, the distribution of confounder
representations on treatment $t$ is equivalent to the counterfactual distribution of confounder representations on treatment $T \neq t$ and thus eliminating the confounding bias.
Additionally, we add the unit's self-confounder $\mathbf{X}_{c}$ as a residual module to emphasize its importance.
Then we can conclude the unit representations for factual prediction $\mathbf{H}_{f}$ and counterfactual prediction $\mathbf{H}_{cf}$,
\begin{equation}
    \mathbf{H}_{f} = \mathbf{E}_{a} + \mathbf{E}_{c} + \mathbf{X}_{c},
\end{equation}
\begin{equation}
    \mathbf{H}_{cf} = \mathbf{E}_{a} + g(\mathbf{X}_{c},\mathbf{E}_{a}) + \mathbf{X}_{c}.
\end{equation}
During the training stage, we only have access to factual observed outcomes. Our first goal is to minimize the error between the inferred outcome and ground truth. To predict the outcome, we utilize a two-branch multilayer perceptron following the classical T-learner \cite{kunzel2019metalearners} architecture, which learns an output function $f: \mathbb{R}^{d} \times \{0,1\}\rightarrow\mathbb{R}$.
\begin{equation}
    f\left(\mathbf{H}_{f,i}, t\right)=\left\{\begin{array}{l}
                                             f_1\left(\mathbf{H}_{f, i}\right) \text { if } t_i=1, \\
                                             f_0\left(\mathbf{H}_{f, i}\right) \text { if } t_i=0.
    \end{array}\right.
\end{equation}
Then the mean squared error is adopted as our factual outcome loss function to minimize the error between the inferred potential outcomes and the ground truth,
\begin{equation}
    \mathcal{L}_{prediction} = \frac{1}{N} \sum_{i=1}^N\left( f\left(\mathbf{H}_{f,i}, t_{i}\right)-y_i\right)^2.
\end{equation}
Besides, we also exploit function $f$ to prediction counterfactual outcome as $f(\mathbf{H}_{cf,i}, 1-t_i)$.
The final loss is defined as,
\begin{equation}
    \begin{aligned}
        \mathcal{L} = &\mathcal{L}_{prediction} + \mathcal{W}_{1}\mathcal{L}_{adjustment} \\&+ \mathcal{W}_{2}\mathcal{L}_{confounder} + \mathcal{W}_{3}\mathcal{L}_{cf\text{-}confounder},
    \end{aligned}
\end{equation}
where $\mathcal{W}_1$, $\mathcal{W}_2$, $\mathcal{W}_3$ are weighted parameters.

\section{Experiment}


\subsection{Experimental Settings}

\subsubsection{Datasets}

Since the lack of ground truth of ITEs is a well-known issue, a common solution is to use artificially simulated data to generate all potential outcomes under different treatments. We adopt two semi-synthetic datasets from \cite{DBLP:conf/wsdm/GuoLL20} as our benchmark datasets both generated from real-world networked data sources.

\textbf{BlogCatalog}~\cite{DBLP:conf/wsdm/GuoLL20}. BlogCatalog dataset is generated from an online community where users plot blogs with 5,196 units and 171,743 edges. Each unit in this dataset is a blogger and each edge indicates a social connection between two bloggers. This dataset adopts the bag-of-words representations of keywords in bloggers' descriptions as each unit's features. For the synthesizing process, the opinions of readers on each blogger are the outcomes and the treatments are denoted by whether contents created by a blogger receive more views on mobile devices or desktops. The treatment $t=1$ or $t=0$ represents the blogger's blogs are read more on mobile devices or on desktop. With the assumption of confounding bias existence, a blogger and his neighbors' topics not only causally influence his treatment assignment but also affect readers' opinions. An LDA topic model on a large set of documents is used as the base model to synthesize treatments and outcomes. Besides, this dataset uses hyper-parameters $\kappa$ to control the magnitude of the confounding bias resulting from neighbors' topics and create three datasets with $\kappa=0.5, 1, 2$. The larger $\kappa$ is, the more significant the influence of neighbors' topics on the device is.

\textbf{Flickr}~\cite{DBLP:conf/wsdm/GuoLL20}. Flickr is a dataset created from a photo-sharing social community with 7,575 units and 12,047 edges, in which each unit is a user and each edge indicates the social relationship between two users. Users' features are represented by a list of tags of interest. Flickr dataset adopts the same settings and simulation procedures as BlogCatalog and also generates three sub-datasets with different magnitudes of the confounding bias.



\begin{table*}[]
    \begin{tabular}{c|cccccc|cccccc}
        \toprule
        & \multicolumn{6}{c|}{BlogCatalog}                                               & \multicolumn{6}{c}{Flickr}                                                    \\ \cmidrule{2-13}
        $\kappa$                & \multicolumn{2}{c}{$\kappa$=0.5} & \multicolumn{2}{c}{$\kappa$=1} & \multicolumn{2}{c|}{$\kappa$=2} & \multicolumn{2}{c}{$\kappa$=0.5} & \multicolumn{2}{c}{$\kappa$=1} & \multicolumn{2}{c}{$\kappa$=2} \\ \cmidrule{2-13}
        Method           & $\sqrt{\epsilon_{PEHE}}$         & $\epsilon_{ATE}$        & $\sqrt{\epsilon_{PEHE}}$        & $\epsilon_{ATE}$       & $\sqrt{\epsilon_{PEHE}}$       & $\epsilon_{ATE}$        & $\sqrt{\epsilon_{PEHE}}$         & $\epsilon_{ATE}$        & $\sqrt{\epsilon_{PEHE}}$        & $\epsilon_{ATE}$       & $\sqrt{\epsilon_{PEHE}}$       & $\epsilon_{ATE}$        \\ \midrule
        BART             & 4.808        & 2.68       & 5.77        & 2.278     & 11.608     & 6.418      & 4.907        & 2.323      & 9.517       & 6.548     & 13.155     & 9.643      \\
        CF               & 7.456        & 1.261      & 7.805       & 1.763     & 19.271     & 4.05       & 8.104        & 1.359      & 14.636      & 3.545     & 26.702     & 4.324      \\
        TARNet           & 11.57        & 4.228      & 13.561      & 8.17      & 34.42      & 13.122     & 14.329       & 3.389      & 28.466      & 5.978     & 55.066     & 13.105     \\
        CFR-mmd          & 11.536       & 4.127      & 12.332      & 5.345     & 34.654     & 13.785     & 13.539       & 3.35       & 27.679      & 5.416     & 53.863     & 12.115     \\
        CFR-Wass         & 10.904       & 4.257      & 11.644      & 5.107     & 34.848     & 13.053     & 13.846       & 3.507      & 27.514      & 5.192     & 53.454     & 13.269     \\
        CEVAE            & 7.481        & 1.279      & 10.387      & 1.998     & 24.215     & 5.566      & 12.099       & 1.732      & 22.496      & 4.415     & 42.985     & 5.393      \\
        \midrule
        NetDeconf        & 4.532        & 0.979      & 4.597       & 0.984     & 9.532      & 2.130      & 4.286        & 0.805      & 5.789       & 1.359     & 9.817      & 2.700      \\
        GIAL            & 4.023        & 0.841      & 4.091       & 0.883     & 8.927      & 1.78       & 3.938        & 0.682      & 5.317       & 1.194     & 9.275      & 2.245      \\ 
         GNUM             & 4.122        & 0.932      & 4.367       & 0.973     & 9.327      & 2.082      & 4.102        & 0.801      & 5.345       & 1.267     & 9.756      & 2.675     \\
         IGL        & 3.987         & 0.815      & 4.012       & 0.845      & 8.746      & 1.628       & 4.012       & 0.703      & 5.437       & 1.259    & 9.312      & 2.321      \\
         \midrule
        DRCFR            & 3.772        & 0.698      & 3.857       & 1.070     & 6.077      & 1.405      & 4.019        & 0.703      & 5.367       & 1.304     & \textbf{7.915}      & 2.691      \\ \midrule
        GDC              & $\textbf{3.012}$        & $\textbf{0.236}$      & $\textbf{3.499}$       & $\textbf{0.591}$     & $\textbf{5.941}$      & $\textbf{0.876}$      & $\textbf{3.952}$        & $\textbf{0.281}$      & $\textbf{5.351}$       & $\textbf{0.491}$     & 8.287      & $\textbf{1.241}$    \\ 
        \bottomrule
    \end{tabular}
    \caption{Overall performance on BlogCatalog and Flickr datasets comparing the effectiveness of GDC and baseline methods.}
    \label{tab: main_results}
\end{table*}

\subsubsection{Baseline Methods}

To investigate the superiority of the proposed
method, we compare it with the following three categories of baselines:
i) Traditional methods (i.e., \textbf{BART}~\cite{hill2011bayesian}, \textbf{CF}~\cite{wager2018estimation}, \textbf{TARNet}~\cite{DBLP:conf/icml/ShalitJS17}, \textbf{CFR-Wass}~\cite{DBLP:conf/icml/ShalitJS17}, \textbf{CFR-MMD}~\cite{DBLP:conf/icml/ShalitJS17}, and \textbf{CEVAE}~\cite{DBLP:conf/nips/LouizosSMSZW17})
ii) Methods considering disentangled representations (i.e., \textbf{PDRCFR}~\cite{DBLP:conf/iclr/HassanpourG20}) and 
iii) Methods considering graph structure (i.e., \textbf{NetDeconf}~\cite{DBLP:conf/wsdm/GuoLL20}, \textbf{GIAL}~\cite{DBLP:conf/kdd/ChuR021}, \textbf{GNUM}~\cite{zhu2023graph}, \textbf{IGL}~\cite{sui2024invariant})

\subsubsection{Evaluation Metrics}

We adopt two widely used evaluation metrics in causal inference~\cite{DBLP:conf/wsdm/GuoLL20,DBLP:conf/kdd/ChuR021}, the Rooted Precision in Estimation of Heterogeneous Effect($\sqrt{\epsilon_{PEHE}}$) \cite{hill2011bayesian} and Mean Absolute Error on ATE($\epsilon_{ATE}$) \cite{imbens2015causal}.

\subsubsection{Implementation}


Firstly, we randomly sample 60\%/20\%/20\% of all units from each dataset as the training/validation/test sets.
Our primary aim is to assess the efficacy of our proposed method on datasets with varying degrees of confounding bias. To achieve this goal, we perform simulation procedures ten times for each dataset and present the average mean of all simulations as the result of each experiment.
To establish baseline results, default settings are employed for all methods. However, it is important to note that all baselines, with the exception of NetDeconf and GIAL, were not initially developed to handle networked observational data. To ensure a fair comparison, we augment other baselines with structural information.
For BART, Causal Forest, TARNet, and CFRNet, the original features of each unit are concatenated with the corresponding row of adjacency matrix and used as the methods' inputs. 
As for the disentangle-based methods DRCFR, an encoder based on graph neural networks (GNN) is first applied to extract encoded feature representations. These representations are then utilized as input for the proposed model. 
We recognize the availability of several alternative graph neural network frameworks (i.e, GCN~\cite{gcn2017}, GraphSAGE~\cite{hamilton2017inductive} and GAT~\cite{gat2018}) for establishing graph-based baselines. Therefore, we will evaluate the performance of these frameworks and report the best results in the subsequent experiments.
Our GDC model employs the original implementation of GAT with two layers, while the hidden units are set to 256. ADAM optimizer\cite{kingma2014adam} is used in our method to minimize the objective function. Following the previous convention, we add a $l_2$ norm regularization with hyperparameter $10^{-4}$ to prevent overfitting. Each training process contains 200 epochs with an initial learning rate of 0.01. A parameter study is also conducted to find the optimal balancing weights $\{\mathcal{W}_1, \mathcal{W}_2, \mathcal{W}_3\}$ in the final loss function.


\subsection{Results Comparison}

The overall performances of different methods on two datasets are demonstrated in Table \ref{tab: main_results} and we summarize several observations: 
As our datasets and experiment settings are aligned to those of the paper \cite{DBLP:conf/kdd/ChuR021}, we obtain partial results of the baselines from that paper.
\begin{itemize}[leftmargin=*]
    \item[$\blacktriangleright$] Our method, GDC, exhibits superior performance in terms of PEHE and ATE on both datasets, compared to other competitive baselines. This improvement in the estimation of Individual Treatment Effects (ITE) underscores the efficacy of our approach, which carefully recognizes the distinct roles that different feature factors play in graph aggregation and skillfully leverage disentangle techniques for performance improvement.
    
    \item[$\blacktriangleright$] Conventional approaches, such as BART, CF, CFR, and CEVAE, display subpar performance due to their inability to integrate auxiliary network information. Although we aim to improve their effectiveness by incorporating the adjacency matrix with the original features, the issue of high dimensionality and sparsity curtails the efficacy of this strategy.
    \item[$\blacktriangleright$] GNN-based methods (i.e., NetDeconf, GIAL, GNUM and IGL)surpass traditional approaches due to the superior representation capacity of GNNs in capturing hidden confounders.
    \item[$\blacktriangleright$] The disentangle-based method, DRCFR, exhibits competitive performance. This approach utilizes a GNN encoder to generate powerful representations and effectively disentangles these representations for practical use. These experimental results underscore the significance of disentangling representations.
    
\end{itemize}

\begin{table}[]
\resizebox{\linewidth}{!} {
    \begin{tabular}{ccccccc}
        \hline
        \multicolumn{7}{c}{BlogCatalog}                                                               \\ \hline
        & \multicolumn{2}{c}{$\kappa$=0.5} & \multicolumn{2}{c}{$\kappa$=1} & \multicolumn{2}{c}{$\kappa$=2} \\ \hline
        Method        & $\sqrt{\epsilon_{PEHE}}$        & $\epsilon_{ATE}$         & $\sqrt{\epsilon_{PEHE}}$       & $\epsilon_{ATE}$        & $\sqrt{\epsilon_{PEHE}}$        & $\epsilon_{ATE}$       \\
        w/o dis       & 4.827       & 0.799       & 4.373      & 0.903      & 9.068       & 2.791     \\
        w/o con       & 3.854       & 0.581       & 3.819      & 0.615      & 6.433       & 1.601     \\
        GDC           & $\textbf{3.012}$       & $\textbf{0.236}$       & $\textbf{3.499}$      & $\textbf{0.591}$      & $\textbf{5.941}$       & $\textbf{0.876}$     \\ \hline \hline
        \multicolumn{7}{c}{Flickr}                                                                    \\ \hline 
        & \multicolumn{2}{c}{$\kappa$=0.5} & \multicolumn{2}{c}{$\kappa$=1} & \multicolumn{2}{c}{$\kappa$=2} \\ \hline
        Method        & $\sqrt{\epsilon_{PEHE}}$        & $\epsilon_{ATE}$         & $\sqrt{\epsilon_{PEHE}}$       & $\epsilon_{ATE}$        & $\sqrt{\epsilon_{PEHE}}$        & $\epsilon_{ATE}$       \\
        w/o dis       & 4.210       & 0.370       & 6.589      & 1.659      & 9.530      & 2.213     \\
        w/o con       & 3.976       & 0.609       & 5.535      & 0.732       & 8.348       & 1.674      \\
        GDC           & $\textbf{3.952}$       & $\textbf{0.281}$       & $\textbf{5.351}$      & $\textbf{0.491}$      & $\textbf{8.287}$       & $\textbf{1.241}$     \\ \hline
    \end{tabular}
    }
    \caption{Experimental results of ablation studies. }
\end{table}

\subsection{Ablation Study}

To evaluate the effectiveness of the design of our framework, we construct various variations based on the current model and conduct ablation studies on both datasets, i.e., \textbf{w/o dis} removing the disentangle module and \textbf{w/o con} keeping the disentangle module but only utilizing adjustment variables.
The experimental results yield several notable observations:



\begin{itemize}[leftmargin=*]
    \item[$\blacktriangleright$] The model lacking the disentangle module exhibits inadequate performance, thereby highlighting the crucial role of the disentangling process in controlling confounding bias.
    \item[$\blacktriangleright$] Our results demonstrate that utilizing only adjustment representations leads to superior performance compared to the without-disentangle variant, which demonstrates the influence of confounding variables. However, these results still fall short of the performance achieved by GDC, showing that relying solely on adjustment variables is inadequate for accurately estimating ITE.

\end{itemize}

\begin{figure}[htb]
    \centering
    \includegraphics[width=0.47\textwidth]{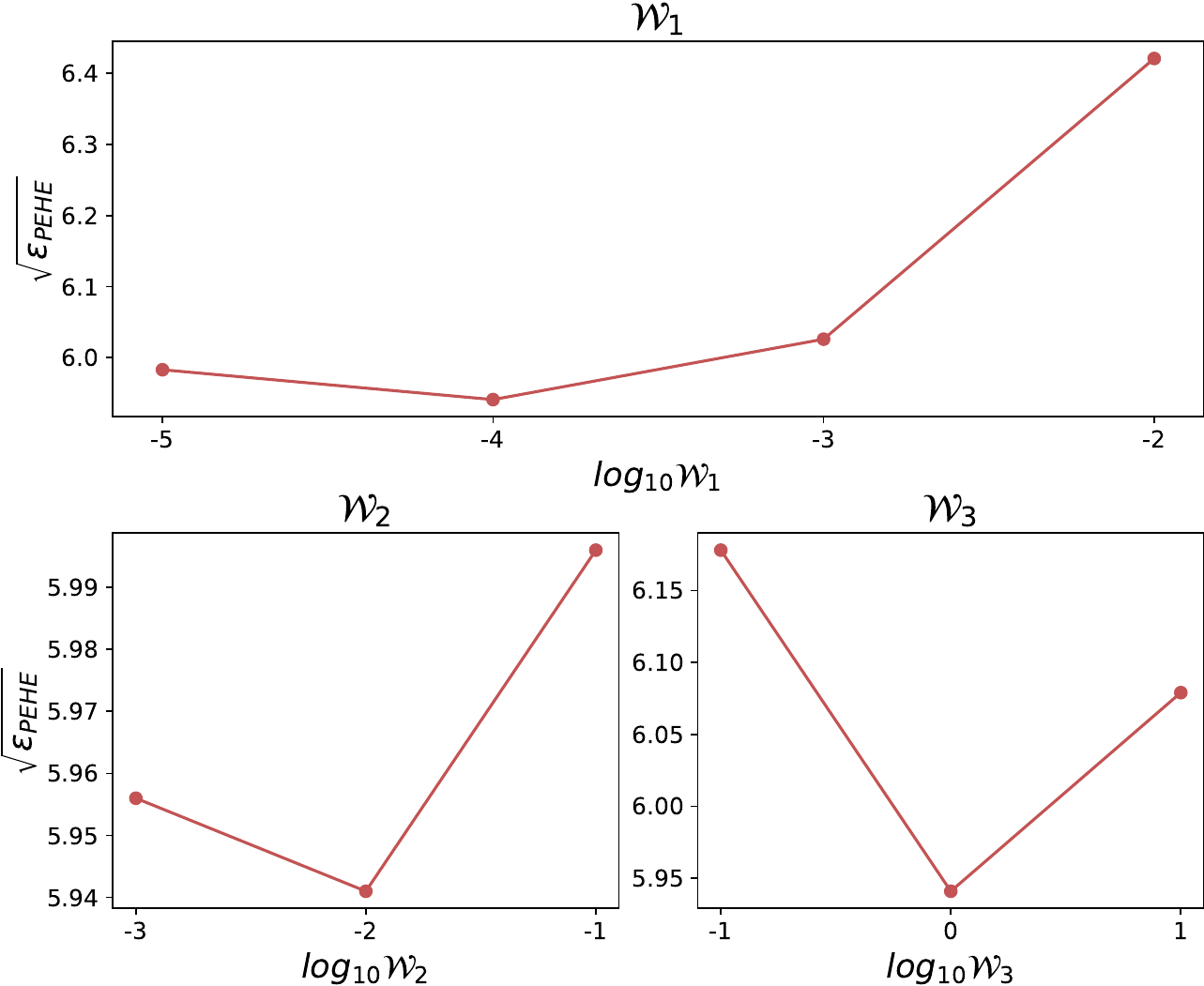}
    \caption{Experimental results of parameter studies. The $\sqrt{\epsilon_{PEHE}}$ 
    initially decreases and then increases as the values of $\mathcal{W}_1$,$\mathcal{W}_2$,$\mathcal{W}_3$ increase.}
    \label{fig: parameter_study}
\end{figure}

\subsection{Hyper-Parameter Study}

In this subsection, we investigate the effect of three hyperparameters on the loss function. Regarding the settings for the parameter studies, we vary $\mathcal{W}_1$ in the range of \{0.00001, 0.0001, 0.001, 0.01\}, $\mathcal{W}_2$ in the range of \{0.001, 0.01, 0.1\} and $\mathcal{W}_3$ in the range of \{0.1, 1, 10\} according to their type and magnitude. Take the BlogCatalog dataset with $\kappa=2$ as an example, the $\sqrt{\epsilon_{PEHE}}$ results of different parameters are shown in Figure~\ref{fig: parameter_study}. We can see that the performance has a trend of rising first and then failing as all three weights increase. The best result occurs when $\mathcal{W}_1 = 0.0001, \mathcal{W}_2 = 0.01, \mathcal{W}_3 = 1$.




\subsection{Visualized Interpretation}

\begin{figure}
    \centering
    \includegraphics[width=0.45\textwidth]{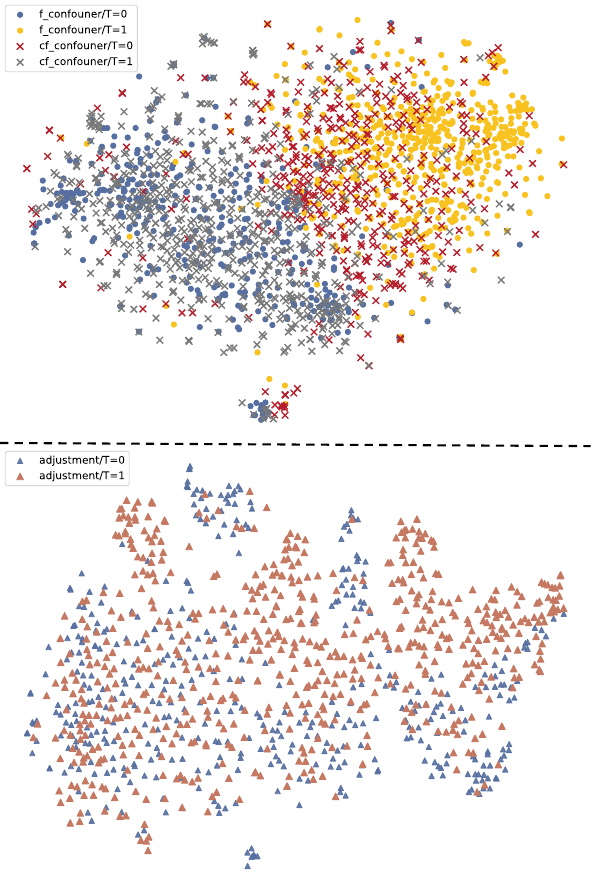}
    \caption{T-SNE projections of learned confounder and adjustment embeddings.}
    \label{fig: embedding_study}
\end{figure}

To further examine whether the learned disentangled embeddings conform to our expectations, we employ t-SNE~\cite{van2008visualizing} to project the learned aggregated embeddings onto a two-dimensional plane in figure~\ref{fig: embedding_study}, using a single realization of BlogCatalog with $\kappa=2$. 

The upper figure illustrates the distribution of confounder and counterfactual confounder across two groups, where different groups are clearly separated, i.e, $p(\mathbf{E}_{c}|T=t)$ and $p(\mathbf{E}_{c}|T\neq t)$ are in different distribution, as indicated by the distinct colors of the circles/crosses. This observation suggests that our method effectively characterizes confounders into distinct groups, maintaining the diversity of confounder representations across different treatments, unlike traditional methods that typically conflate them. Furthermore, the mixed distribution of the confounder from one group and the counterfactual confounder from the opposing group, i.e., $p(\mathbf{E}_{c}|T=t)$ and $p(\mathbf{E}_{cf}|T\neq t)$ share the same distribution, indicates that our model is capable of utilizing counterfactual confounder representations to approximate the confounder representation with an opposing treatment and thus our model has eliminated confounding bias.
In the below figure, we observe a mixed distribution of the adjustment across two groups.
This observation provides further confirmation that our model effectively captures and represents the adjustment factor which is independent of the treatment.

\section{RELATED WORK}

In this section, we introduce several related works.


Causal inference from observational data has been a critical research topic since its early days. Bayesian Additive Regression Trees (BART)~\cite{hill2011bayesian} and Causal Forest~\cite{wager2018estimation} were among the early tree-based methods that played a significant role. With the emergence of powerful neural networks, a series of popular methods have been proposed~\cite{DBLP:conf/icml/ShalitJS17,DBLP:conf/icml/JohanssonSS16}. For example, Counterfactual Regression~\cite{DBLP:conf/icml/ShalitJS17} utilizes deep neural networks to learn complex representations and employs Integral Probability Metrics regularization to control confounding bias. DRCFR~\cite{DBLP:conf/ijcai/HassanpourG19} argues that previous methods should not remove all discrepancies between confounders of different groups and proposes disentangle-based methods leveraging different parts of features while retaining the predictive power of the confounder. 


GNNs have proven to be a powerful tool for handling graph data~\cite{wang2024optimizing,hu2022merit}, and their potential for promoting development has been recognized across various research areas, including causal inference. NetDeconf~\cite{DBLP:conf/wsdm/GuoLL20} is a pioneering work that harnesses GCN for causal inference with observational data, exploiting it to capture network information and identify a proxy of hidden confounders. GIAL~\cite{DBLP:conf/kdd/ChuR021} builds upon this idea and highlights the particularity of graph data in causal inference, noting the imbalanced network structure that distinguishes it from traditional graph learning tasks and recommending the use of a mutual information regularizer to address this issue. More recently, several initiatives have emerged that integrate GNNs with causal inference. However, these efforts tend to focus on specific scenarios, such as situations with limited labels or the need to capture spillover effects.  Our work builds upon their insights and takes a step further, recognizing the distinct roles that different feature factors play in graph aggregation and leveraging disentangle techniques to address this challenge.

\section{Conclusion}



In the paper, we introduce a novel framework called the Graph Disentangled Causal Model (GDC) to tackle the challenge of estimating individualized treatment effects using networked observational data. 
This model categorizes each unit's attributes into adjustment factors and confounder factors with  disentangled representation learning.
We implement three distinct aggregation methods based on the causal graph of the network data to derive aggregated adjustment, confounder, and counterfactual confounder factors. 
By synthesizing these aggregated factors, GDC effectively estimates both factual and counterfactual outcomes. Extensive experiments conducted on two semi-synthetic datasets validate the efficacy of our approach.

\bibliographystyle{ACM-Reference-Format}
\balance
\bibliography{reference}



\end{document}